\documentclass[lettersize,journal]{IEEEtran}
\usepackage{amsmath,amsfonts}
\usepackage{algorithmic}
\usepackage{algorithm}
\usepackage{array}
\usepackage[caption=false,font=normalsize,labelfont=sf,textfont=sf]{subfig}
\usepackage{textcomp}
\usepackage{stfloats}
\usepackage{url}
\usepackage{verbatim}
\usepackage{graphicx}
\usepackage{multirow}
\usepackage{booktabs}
\usepackage{siunitx}
\usepackage{xcolor}
\usepackage{romannum} 
\usepackage{cite}
\usepackage{tabularx}
\usepackage{ragged2e} 
\begin{document}

\title{Memory-DD: A Low-Complexity Dendrite-Inspired Neuron \\for Temporal Prediction Tasks}

\author{Dongjian Yang, Xiaoyuan Li, Chuanmei Xi, Ye Sun, Gang Liu* 

\thanks{*Corresponding authors: Gang Liu(gangliu\_@zzu.edu.cn)}
\thanks{This work is supported by the National Natural Science Foundation of China (62303423), the STI 2030-Major Project(2022ZD0208500), Postdoctoral Science Foundation of China (2024T170844,2023M733245), the Henan Province key research and development and promotion of special projects (242102311239),Shaanxi Province key research and development plan(2023GXLH-012)}

\thanks{}}

\markboth{}%
{Shell \MakeLowercase{\textit{et al.}}: A Sample Article Using IEEEtran.cls for IEEE Journals}


\maketitle

\begin{abstract}
Dendrite-inspired neurons have been widely used in tasks such as image classification due to low computational complexity and fast inference speed. Temporal data prediction, as a key machine learning task, plays a key role in real-time scenarios such as sensor data analysis, financial forecasting, and urban traffic management. However, existing dendrite-inspired neurons are mainly designed for static data. Studies on capturing dynamic features and modeling long-term dependencies in temporal sequences remain limited. Efficient architectures specifically designed for temporal sequence prediction are still lacking. In this paper, we propose Memory-DD, a low-complexity dendrite-inspired neuron model. Memory-DD consists of two dendrite-inspired neuron groups that contain no nonlinear activation functions but can still realize nonlinear mappings. Compared with traditional neurons without dendritic functions, Memory-DD requires only two neuron groups to extract logical relationships between features in input sequences. This design effectively captures temporal dependencies and is suitable for both classification and regression tasks on sequence data. Experimental results show that Memory-DD achieves an average accuracy of 89.41\% on 18 temporal classification benchmark datasets, outperforming LSTM by 4.25\%. On 9 temporal regression datasets, it reaches comparable performance to LSTM, while using only 50\% of the parameters and reducing computational complexity (FLOPs) by 27.7\%. These results demonstrate that Memory-DD successfully extends the low-complexity advantages of dendrite-inspired neurons to temporal prediction, providing a low-complexity and efficient solution for time-series data processing.
\end{abstract}

\begin{IEEEkeywords}
Dendrite-inspired neuron, temporal prediction, low-complexity neuron.
\end{IEEEkeywords}

\section{Introduction}
\IEEEPARstart{W}{ith} the rapid development of information technology, massive temporal sequence data have become the foundation of modern society, ranging from industrial IoT to financial markets. Accurate and fast prediction of such data is a key prerequisite for higher-level applications, including efficient decision support, real-time anomaly detection, and optimized resource allocation. In latency-sensitive scenarios, the high-speed processing ability of temporal prediction models is no longer optional but has become a decisive factor for their practical success.

To address the challenges of temporal prediction tasks, deep learning methods have become the mainstream approach. Recurrent neural networks (RNNs) \cite{ref1}, represented by long short-term memory (LSTM) \cite{ref2}, capture temporal dynamics through gating mechanisms. Transformer architectures leverage self-attention and have made breakthroughs in long-range dependency modeling \cite{ref3}. However, the strong performance of these models often comes at the cost of heavy computation, with large parameter sizes and intensive resource demands. The sequential processing bottleneck of RNNs and the quadratic complexity of Transformers both impose serious limitations in latency-critical scenarios \cite{ref4}. To address these challenges, there is an urgent need to develop a lightweight temporal prediction model that can achieve high accuracy with low complexity. 

In the search for models with low complexity and biological plausibility, spiking neural networks (SNNs), known as the third generation of neural networks, have attracted wide attention \cite{ref5}. SNNs transmit and process information by simulating the discrete spikes of biological neurons, which gives them natural advantages in handling temporal information and potential for low power consumption. However, the unique event-driven and spike-firing mechanisms of SNNs make it difficult to directly apply mature gradient-based optimization algorithms. This leads to more complex training procedures and heavy computational costs, which greatly limit their applications to broader and more complex problems \cite{ref6}.In view of this, some research has shifted from simulating the macroscopic firing behavior of neurons to exploring more microscopic computational principles within neurons. The goal is to find an alternative path that combines biological inspiration with computational efficiency. Among these efforts, dendrite-inspired neuron models represent a highly promising direction. Biological studies have shown that dendrites are not passive signal collectors. Instead, they can actively perform complex local nonlinear operations, such as AND, OR, and XOR \cite{ref7,ref8,ref9,ref10}. Inspired by this, our group previously proposed dendrite-inspired models such as Dendrite Net (DD) \cite{ref11,ref12}. These models use efficient operations, such as the Hadamard product, to simulate dendritic logical interactions. In this way, they achieve strong nonlinear mapping capability without relying on complex nonlinear activation functions.Due to their low computational complexity, simple and transparent structures, such models have achieved significant success in static data processing tasks such as image classification. However, existing dendrite-inspired neurons have mainly focused on static data. Their architectures do not contain memory mechanisms and thus cannot directly handle sequential data with temporal dependencies. As a result, the field of temporal prediction still lacks a dendrite-inspired architecture specifically designed for this purpose, one that can inherit the advantages of low complexity while enabling sequence modeling.

This paper proposes Memory-DD, a low-complexity dendrite-inspired neuron model. The model consists of two groups of dendrite-inspired neurons that contain no nonlinear activation functions but can still achieve nonlinear mappings. Compared with traditional neurons without dendritic functions, Memory-DD requires only two neuron groups to extract the logical relationships among features in input sequences. This design effectively captures temporal dependencies and enables accurate modeling of sequence dynamics. Memory-DD is suitable for both classification and regression tasks in temporal prediction.

The contributions of this paper are as follows:

1) We propose Memory-DD, a low-complexity dendrite-inspired neuron model. For the first time, we combine a memory mechanism with a dendrite-inspired structure in an innovative way to address the problem of temporal sequence prediction.

  2) We construct Memory-DD using two groups of dendrite-inspired neurons and demonstrate its effectiveness. Experimental results show that Memory-DD achieves comparable or even better predictive performance than traditional LSTM models, while using only half the number of parameters. This provides a new feasible path for lightweight temporal modeling.

The remainder of this paper is organized as follows. Section \Romannum{2} reviews related work on dendrite-inspired neurons and temporal prediction methods. Section \Romannum{3} presents the overall architecture of Memory-DD, including the biologically inspired neuron design, the core mechanism of the Memory-DD unit, and the complete model construction. Section \Romannum{4} provides a detailed theoretical analysis of computational complexity.Section \Romannum{5} reports comprehensive experimental results, including experimental setup, benchmark datasets, evaluation metrics, performance comparison with mainstream models on temporal classification and regression tasks, and ablation studies. Section \Romannum{6} discusses the theoretical implications and practical value of the results. Section \Romannum{7} concludes the main contributions and innovations of this work and outlines directions for future research.

\section{Related work}
\subsection{Dendrite-Inspired Neuron}
Traditional artificial neural networks, such as multilayer perceptrons (MLPs), abstract neurons as simplified point models \cite{ref13}. Their functions are usually represented by $f(wx+b)$. In essence, this model reduces the complex functions of biological neurons to a linear weighted summation of inputs followed by a single nonlinear activation. However, extensive neuroscience studies have shown that this abstraction overlooks a critical computational structure in biological neurons—the dendrite \cite{ref7,ref8,ref9,ref10}. Dendrites are not merely passive signal transmission channels. Instead, they act as active computational subunits that can perform complex nonlinear integration before synaptic signals reach the soma. Pioneering studies have confirmed that dendritic branches of cortical neurons can implement logical operations such as XOR through mechanisms like localized dendritic action potentials \cite{ref14}. This endows a single neuron with computational power and information-processing efficiency far beyond traditional artificial neuron models. Such dendritic computations and nonlinear integration are much closer to the true feature integration mechanisms of biological neurons.

Based on these biological findings, researchers have developed various dendrite-inspired neuron models to bring the computational advantages of dendrites into machine learning tasks. The core design idea of these models is to replace the single linear weighted summation of traditional neurons with a layered, tree-like structure. A typical dendrite-inspired neuron model contains multiple dendritic branches, and each branch may have sub-branches, forming a hierarchical computational structure. Input signals are first processed locally with nonlinear functions (such as Sigmoid or ReLU) on each dendritic branch. The outputs of these branches are then progressively integrated and finally combined at the soma for global decision-making.For example, Dmixnet, a dendritic multilayer perceptron architecture, was designed for image recognition tasks \cite{ref15}. Another study proposed a dendritic neural network (dANN), which also demonstrated higher accuracy and robustness in image classification \cite{ref16}.The main advantage of this structure lies in its inherent low complexity and efficient feature combination capability. Since a single dendrite-inspired neuron already contains multi-level nonlinear computation within itself, it can learn and map complex feature combinations using far fewer neurons than a traditional MLP. This makes it possible to build more compact and efficient neural network models.

However, current mainstream dendrite-inspired models also have clear limitations. Their architectures are mostly feedforward, which makes them well suited for static data such as image classification. But they generally lack built-in memory mechanisms or recurrent connections. As a result, their ability to model temporal information is limited, and they are difficult to apply directly to dynamic sequence tasks such as video analysis or natural language processing.In addition, although these models show great conceptual potential, their performance validation and standardized comparisons on large-scale public benchmark datasets (such as ImageNet) are still insufficient compared with mainstream convolutional neural networks (e.g., ResNet and EfficientNet) \cite{ref17,ref18}. Research on large-scale applications and training optimization of dendrite-inspired models is still at an early stage.

\subsection{Temporal Prediction Methods}
In the early stage of temporal prediction, traditional statistical methods such as ARIMA and exponential smoothing dominated the field. The ARIMA model, proposed by Box and Jenkins, captures linear relationships in time series by combining three core components: autoregression (AR), differencing (I), and moving average (MA) \cite{ref19}. Exponential smoothing, on the other hand, predicts future values by assigning exponentially decaying weights to past observations \cite{ref20}. These methods were widely favored due to their solid theoretical foundation, simple structure, and ease of implementation. However, they are essentially linear models and perform poorly when facing complex nonlinear patterns, multivariate couplings, and long-term dependencies that are common in real-world data \cite{ref19}. In addition, the strong stationarity assumption of ARIMA is difficult to satisfy in practical applications.

To overcome the linear limitations of traditional statistical models, researchers began to explore deep learning approaches. Long short-term memory networks (LSTM) introduced cell states and gating mechanisms, effectively addressing the gradient vanishing problem of standard RNNs and capturing dependencies spanning hundreds of time steps \cite{ref2}. The gated recurrent unit (GRU), as a simplified variant of LSTM, reduced the number of parameters while achieving comparable or even superior performance in many tasks \cite{ref21}. Despite the remarkable success of LSTM and its variants in temporal modeling, their inherent recurrent structures require sequential computation step by step. This greatly limits their parallelization ability and leads to low training efficiency when handling long sequences.

To completely eliminate the sequential dependency of recurrent neural networks, the Transformer model was introduced \cite{ref3}. Its core self-attention mechanism allows the model to directly compute dependencies between any two time points in a sequence, regardless of distance. This enables efficient parallel processing and powerful long-term dependency modeling. With this "global view," Transformers perform exceptionally well in long-sequence prediction tasks. However, the strength of standard Transformers comes with a critical drawback: the computational complexity of self-attention grows quadratically with the sequence length ($O(L^2)$). As the sequence becomes longer, the computational cost rises sharply, which severely restricts their use in resource-constrained environments.

To address the high complexity of Transformers, researchers have explored a wide range of lightweight temporal prediction models. These approaches mainly reduce computational cost through sparse attention \cite{ref22}, linearized attention \cite{ref23}, or by combining convolutional neural networks \cite{ref24}. Such methods alleviate the quadratic complexity problem to some extent and achieve a balance between performance and efficiency. Nevertheless, most of the existing lightweight approaches remain incremental improvements to traditional deep learning architectures, lacking fundamentally innovative ideas.

\subsection{Existing Research and Positioning of This Work}
In summary, existing research in dendrite-inspired neurons and time series forecasting shows a clear technological separation. On the one hand, as discussed in Section 2.1, dendrite-inspired neurons, motivated by biological findings, have shown great success as a low-complexity and high-efficiency computing paradigm in static data tasks such as image classification. By mimicking the nonlinear integration mechanisms of dendrites, they achieve an excellent balance between parameter efficiency and model performance. However, their inherent feedforward and stateless architecture exposes fundamental limitations when applied to dynamic sequential data. This leaves a clear gap for their use in time series forecasting. On the other hand, as discussed in Section 2.2, mainstream time series forecasting methods—whether LSTM-based models with recurrent mechanisms or Transformer architectures with self-attention—demonstrate strong ability to capture temporal dependencies. Yet, both face the bottleneck of high computational complexity. Although the research community has explored lightweight improvements, these efforts mainly focus on optimizing existing macro-architectures rather than seeking breakthroughs at the level of the neuron computation paradigm itself.

This study lies at the intersection of these two research paths and aims to bridge this gap. We explicitly position the proposed Memory-DD model as a key extension of dendrite-inspired neurons, bringing them for the first time from static data applications into dynamic time series modeling. By innovatively combining memory mechanisms with the efficient computation core of dendrite-inspired neurons, we aim to test a central scientific question: \textbf{Can a novel dendrite-inspired neuron model for time series forecasting be constructed that achieves prediction accuracy comparable to, or better than, traditional LSTM-based models, while consuming fewer computational resources?} Through direct experimental comparisons with LSTM, GRU, and other mainstream models, this paper systematically evaluates the feasibility and advantages of this new low-complexity pathway for time series modeling.

\section{Method}
This section provides a systematic description of the architecture of the proposed dendrite-inspired neuron model (Memory-DD). First, we trace the biological inspiration of the model and introduce its core computational paradigm—the dendrite-inspired neuron.Next, we explain how this paradigm is extended into a dynamic computational unit with memory and recurrent capability.Finally, we describe how these units are organized into a complete Memory-DD model that can process time series data of arbitrary length.

\begin{figure*}[!t]
\centering
\includegraphics[width=0.88\textwidth]{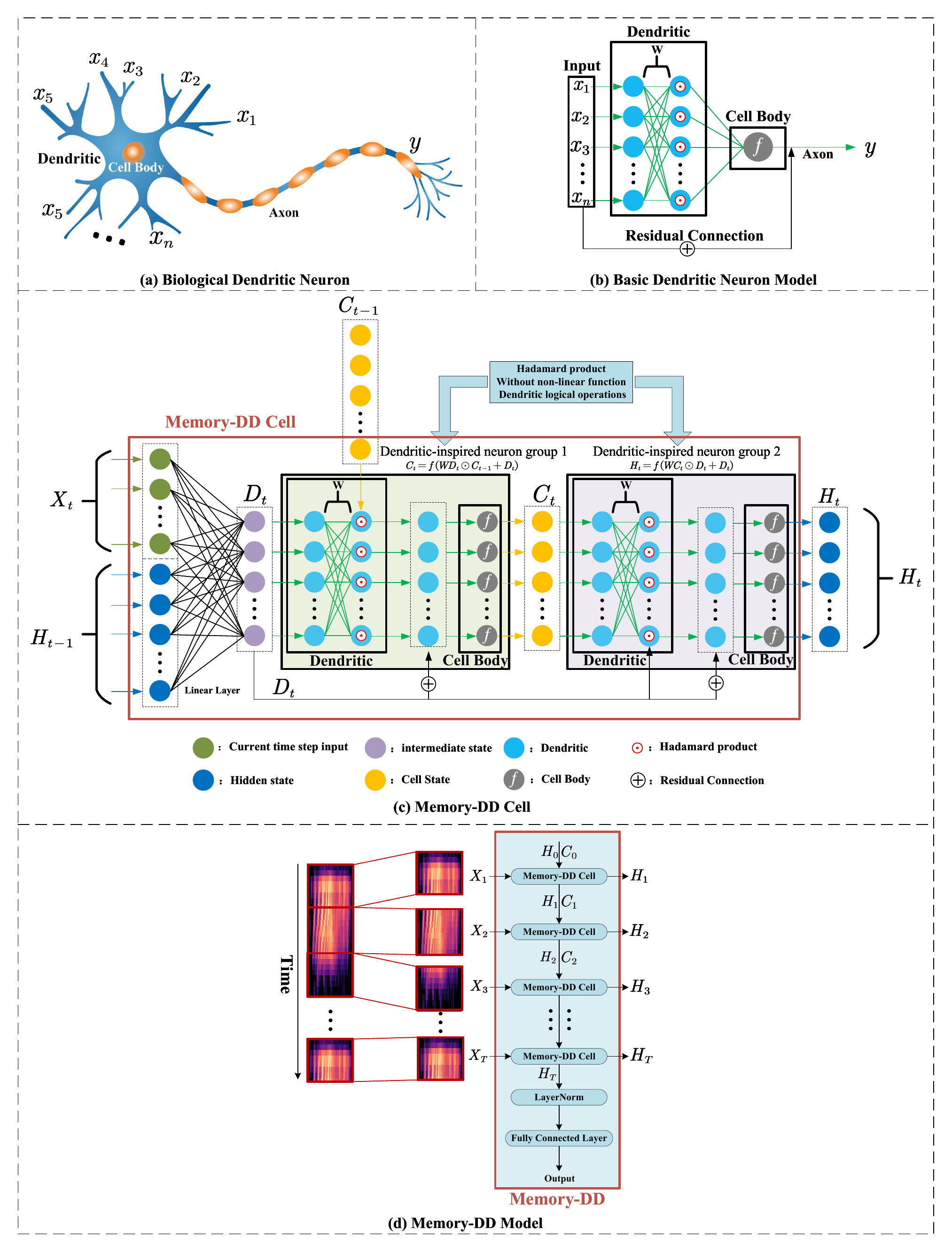}
\caption{The overall architecture of the proposed Memory-DD model. 
(a) The biological neuron, which provides the biological inspiration for our model by integrating signals through its dendrites. 
(b) The architecture of a basic Dendritic Neuron model, which utilizes a dendritic layer to perform non-linear computation via the Hadamard product and incorporates a residual connection. 
(c) The detailed structure of the core Memory-DD Cell, which is composed of a linear layer and two primary functional modules. Dendritic-inspired neuron group 1 serves as the memory update module, updating the cell state to $C_t$ by performing a dendritic logical operation (Hadamard product) between the previous cell state $C_{t-1}$ and an intermediate state $D_t$ (generated from the current input $X_t$ and the previous hidden state $H_{t-1}$). Dendritic-inspired neuron group 2 acts as the decision-making module, generating the new hidden state $H_t$ based on the updated cell state $C_t$ and the intermediate state $D_t$. 
(d) The unrolled architecture of the complete Memory-DD model for processing sequential data. The model processes an input sequence step-by-step, and the final hidden state is passed through a Layer Normalization and a fully connected layer to produce the final prediction.}
\label{fig_1}
\end{figure*}

\subsection{Biological Inspiration from Dendrites: Dendrite-Inspired Neuron Model}
Classical artificial neural networks are largely built on the McCulloch-Pitts "point neuron" model \cite{ref25}. This model abstracts a neuron as a simple linear integrator followed by a single nonlinear activation process. Although this simplification has achieved great success in history, it ignores a key aspect of biological computation: the active role of dendrites. Strong evidence from neuroscience shows that dendrites are not passive signal transmission pathways but complex computational subunits(see Fig. 1(a)). The dendritic branches of a single neuron can perform local nonlinear operations on thousands of synaptic inputs before integration at the soma. One important mechanism is the multiplicative interaction between signals.

Inspired by this, we designed the core computational paradigm of the dendrite-inspired neuron model, as shown in Fig. 1(b). Biological dendrites are capable of performing nonlinear computations, and traditional approaches to mimicking this functionality typically introduce standard activation functions (such as Sigmoid or ReLU) to achieve nonlinearity. However, these conventional methods significantly increase the computational complexity of dendritic operations, which contradicts the inherent computational efficiency observed in biological dendrites. To address this fundamental mismatch, we innovatively propose using the Hadamard product (element-wise multiplication, denoted as $\odot$) as the basic operator for nonlinear interaction. This paradigm aims to directly simulate the multiplicative gating mechanism of dendrites while maintaining computational efficiency. The Hadamard product operation, which modulates one vector by another through element-wise multiplication, provides a direct and computationally efficient simulation of the gating effect among biological dendritic branches, preserving the rapid computational characteristics that make biological dendrites so effective.

\subsection{Memory-DD Cell: A Dendrite-Inspired Recurrent Cell}
To process time-series data, a static dendrite-inspired computation paradigm is not sufficient. We must embed it into a dynamic structure that can maintain internal states and evolve over time. For this purpose, we design the Memory-DD Cell, as shown in Fig. 1(c). This is a dendrite-inspired recurrent Cell that is simple but powerful. Inside the Cell, two Dendritic-inspired Neuron Groups serve as concrete realizations of the dendrite-inspired computation paradigm. They simulate dendritic gating logic to perform memory updates and decision outputs.
\subsubsection{Input Fusion Layer (Context Information Encoding)}

The first stage aims to integrate the current input with the historical context stored in the previous hidden state. Through a linear transformation, the concatenated input is projected into a unified intermediate state $D_t$, which serves as the central information hub for the current time step.
\begin{equation}
D_t = W_1 [H_{t-1}, X_t] + b
\end{equation}

Here, $D_t$ denotes the intermediate state vector generated at time step $t$. $H_{t-1}$ an $X_t$ represent the previous hidden state and the current input, respectively. $[,]$ indicates the vector concatenation operation. $W_1$ and $b$ are the learnable weight matrix and bias vector of this layer.

\subsubsection{Dendritic-inspired Neuron Group 1 (Memory Update Module)}

This module forms the core of the model's long-term memory mechanism. It dynamically filters and updates historical memory through a dendritic-inspired neuron group.
\begin{equation}
C_t = f(WD_t \odot C_{t-1} + D_t)
\end{equation}

Here, $C_t$ denotes the updated cell state at time step $t$.$C_{t-1}$ is the cell state from the previous time step, and $D_t$ is the current contextual information. $W$ is a key learnable weight matrix.$\odot$ represents the Hadamard product, and $f$ is the tanh activation function.

This computation consists of two components. The core is the Hadamard product $WD_t\odot C_{t-1}$, which directly simulates the multiplicative gating computation of dendrites. The gating vector $WD_t$, derived from the current context $D_t$, modulates the historical information $C_{t-1}$ element-wise, enabling dynamic selection of old memory—i.e., selectively "forgetting" or "enhancing" specific dimensions of information. For example, assume that a dimension in the previous memory state $C_{t-1}$ has a value of 0.7. If the corresponding dimension of the gating vector $WD_t$, generated by the current context, has a value of 1.2, then through the Hadamard product this memory dimension is updated to 1.2 × 0.7 = 0.84. This means that the current information strengthens this dimension of long-term memory. If the value of $WD_t$ for that dimension is 0.5, the memory is updated to 0.5 × 0.7 = 0.35, indicating that the current information weakens this memory. When the value of $WD_t$ for that dimension is close to zero (e.g., 0.001), the memory is updated to 0.001 × 0.7 = 0.0007, which is nearly zero. This results in forgetting that dimension of memory.In addition, the addition term $+D_t$ forms a shortcut path. This path serves to supplement and strengthen the direct influence of current information on memory updates. It injects the full, ungated current context $D_t$ directly into the update process, ensuring that new knowledge can be incorporated into long-term memory quickly and without loss.

\subsubsection{Dendritic-inspired Neuron Group 2 (Decision Output Module)}

The final module is responsible for generating the hidden state $H_t$. Here, the second dendritic-inspired neuron group uses the updated long-term memory to gate the current information and form the final decision.
\begin{equation}
H_t = f(WC_t \odot D_t + D_t)
\end{equation}

Here, $H_t$ is the hidden state at the current time step $C_t$ is the updated cell state, and $D_t$ is the current contextual information. $W$ is a shared weight matrix identical to that in the memory update module.$\odot$ represents the Hadamard product, and $f$ is the tanh activation function.

Symmetric to the structure of the memory update module, the dendritic computation $WC_t\odot D_t$ is also the core here. However, its role is transformed: the gating vector $WC_t$, generated from the updated long-term memory $C_t$, now acts as a "filter" to select the current contextual information $D_t$, determining which information is most relevant to the current decision. For example, if a dimension in the gating vector $WC_t$, which represents long-term experience, has a value of 1.5, and the corresponding current information $D_t$ is 0.8, their Hadamard product is 1.5 × 0.8 = 1.2. This indicates that long-term experience amplifies this part of the current information in the final decision.In contrast, if a dimension in $WC_t$ has a value of 0.1, the same current information (0.8) will be suppressed to 0.1 × 0.8 = 0.08, suggesting that its importance is reduced based on past experience.When a dimension in $WC_t$ is close to zero (e.g., 0.001), the corresponding current information (0.8) becomes almost negligible after the Hadamard product, resulting in 0.001 × 0.8 = 0.0008. This means that this dimension of information is ignored in the final decision.This process simulates a decision mechanism where long-term experience weights and selects short-term information.Similarly, the shortcut path $+D_t$ plays a crucial supplementary role. It ensures that the final output $H_t$ contains not only the memory-filtered information but also the original, complete current context. This design can be seen as a form of residual connection, which facilitates gradient propagation. At the same time, it allows the hidden state output to become a richer representation, organically combining the original short-term information with the long-term memory-regulated information. 

\subsection{Memory-DD Model}
With the Memory-DD Cell capable of processing a single time step and passing states, constructing an end-to-end model that handles complete sequences becomes straightforward. The full Memory-DD model is a recurrent neural network formed by unfolding multiple Memory-DD Cells along the temporal axis, as illustrated in Fig. 1(d). The information processing flow in the model is as follows: at time step t = 0, the initial hidden state $H_0$ and cell state $C_0$ are initialized (usually as zero vectors); for each time step $t$ from 1 to $T$, a Memory-DD Cell processes $X_t$, $H_{t-1}$, and $C_{t-1}$, generating $H_t$ and $C_t$, which are passed to the next time step; after processing the entire sequence, the final hidden state $H_T$ represents a global encoding of the whole input sequence. This final state is then fed into a Layer Normalization layer to stabilize the learned representation, followed by a fully connected layer that maps it to the target prediction space (e.g., class probabilities for classification or continuous values for regression).

In summary, the architecture of Memory-DD follows a clear logical chain: it extracts a Hadamard product-based computational paradigm from the multiplicative gating principle of biological dendrites; this paradigm is instantiated into two functionally distinct and complementary dendritic-inspired neuron groups, encapsulated into a recurrent cell with tightly coupled memory and decision mechanisms; finally, a complete time-series model is built that is powerful in representation yet extremely compact in computation and parameter requirements.

\section{Computational Complexity Analysis}
To quantitatively evaluate the computational efficiency of the Memory-DD model, this section provides a detailed derivation of its parameter count and floating-point operations(FLOPs), followed by a comparative analysis with mainstream time-series prediction models. Let the hidden state dimension of the Memory-DD model be $d_h$ and the input feature dimension be $d_x$. According to the architecture described in Section 3.1, Memory-DD includes the following computational steps when processing L time steps:

Step 1: Input Fusion Layer
\begin{equation}
D_t = W_1 [H_{t-1}, X_t] + b
\end{equation}

In this step, $W_1 \in \mathbb{R}^{d_h \times (d_x + d_h)}$, $b \in \mathbb{R}^{d_h}$. The number of parameters is $d_h(d_x + d_h + 1)$, and the FLOPs are $L(2d_h d_x + 2d_h^2)$.
.

Step 2: Dendritic-inspired Neuron Group 1: Memory Update Module
\begin{equation}
C_t = \tanh(WD_t \odot C_{t-1} + D_t)
\end{equation}

Here, $W \in \mathbb{R}^{d_h \times d_h}$, and $\odot$ denotes the Hadamard product. The number of parameters is $d_h^2$, and the FLOPs are $L(2d_h^2 + d_h)$.

Step 3: Dendritic-inspired Neuron Group 2: Decision Output Module
\begin{equation}
H_t = \tanh(WC_t \odot D_t + D_t)
\end{equation}

The same weight matrix $W$ as in Step 2 is used (weight sharing). The number of parameters is $d_h^2$, and the FLOPs are $L(2d_h^2 + d_h)$. Therefore, the total complexity of Memory-DD is:
Total number of parameters: $2d_h^2 + d_h d_x + d_h$; FLOPs over L time steps: $L(6d_h^2 + 2d_h d_x + 2d_h)$.

To comprehensively evaluate the computational efficiency of Memory-DD, we compare it with five mainstream time-series prediction models. LSTM manages complex memory through three gated units (forget gate, input gate and output gate), each requiring a separate weight matrix. GRU, as a simplified version of LSTM, uses two gates (reset gate and update gate). BiLSTM captures bidirectional temporal dependencies using forward and backward LSTM networks. TCN builds a deep network using causal convolutions and residual connections. Transformer relies on multi-head self-attention and feedforward layers.

Table \Romannum{1} summarizes the theoretical computational complexity of these models in terms of parameter count and FLOPs over L time steps. By formally comparing the number of parameters and FLOPs for each mainstream model, the fundamental design advantage of Memory-DD becomes clear. Traditional recurrent neural networks, such as LSTM and GRU, have computational complexity dominated by the large matrix multiplications required for their multiple internal gates (three and two, respectively). This makes both their parameter count and FLOPs scale with the square of the hidden dimension $d_h$, with relatively large coefficients (4 and 3, respectively). Memory-DD replaces the redundant gating structures of traditional RNNs with two highly compact dendritic-inspired neuron groups and a weight-sharing mechanism. This reduces computational overhead to a minimum while preserving the core representational power—$d_h^2$ complexity—and even achieves fewer parameters than the already lightweight GRU.When compared to more complex architectures such as BiLSTM, TCN, and Transformer, the advantage of Memory-DD becomes even more pronounced. For example, the Transformer not only possesses a substantially larger parameter count (with its feedforward network dimension $d_{ff}$ typically much greater than $d_h$), but its FLOPs complexity $2Ld_xd_h + 4L^2d_h + 8Ld_h^2 + 4Ld_hd_{ff}$ includes terms quadratic in the sequence length $L$, resulting in computational costs that escalate quadratically for longer sequences. In summary, through a fundamental innovation in the computational paradigm, Memory-DD theoretically ensures the lowest computational complexity and parameter scale among all compared models.

\begin{table*}[!t]
\centering
\caption{Theoretical computational complexity comparison of different models. In TCN, $k$ is the convolution kernel size and $\text{layers}$ is the number of network layers; in Transformer, $d_{\text{ff}}$ is the hidden dimension of the feedforward network; $L$ is the sequence length.}
\label{tab:complexity_comparison}

\renewcommand{\arraystretch}{1.5}

\begin{tabular}{lcc}
\toprule  
Models & Parameters & FLOPs \\
\midrule  
LSTM & $4d_h^2 + 4d_h d_x + 4d_h$ & $L(8d_h^2 + 8d_h d_x + 4d_h)$ \\
GRU & $3d_h^2 + 3d_h d_x + 3d_h$ & $L(6d_h^2 + 6d_h d_x + 2d_h)$ \\
BiLSTM & $8d_h^2 + 8d_h d_x + 8d_h$ & $L(16d_h^2 + 16d_h d_x + 8d_h)$ \\
TCN & $d_x d_h + k d_h^2 \cdot layers$ & $2L(d_x d_h + k d_h^2 \cdot layers)$ \\
Transformer & $d_x d_h + 4d_h^2 + 2d_h d_{ff}$ & $2L d_x d_h + 4L^2 d_h + 8L d_h^2 + 4L d_h d_{ff}$ \\
\textbf{Memory-DD (Ours)} & $\boldsymbol{2d_h^2 + d_h d_x + d_h}$ & $\boldsymbol{L(6d_h^2 + 2d_h d_x + 2d_h)}$ \\
\bottomrule  
\end{tabular}
\end{table*}

The core mechanisms by which Memory-DD achieves significant computational efficiency improvements are as follows:

1) Weight sharing strategy significantly reduces parameter count: The two dendritic-inspired neuron groups in Memory-DD share the same weight matrix, avoiding the parameter redundancy caused by multiple independent gating matrices in traditional models. Compared with LSTM, which requires four independent weight matrices, TCN with multiple convolutional layers, and Transformer with multi-head attention and feedforward network parameters, Memory-DD only requires a single shared matrix, achieving substantial parameter compression.

2) Hadamard product provides efficient nonlinear interactions: Memory-DD uses the Hadamard product to realize nonlinear gated interactions between features. This element-wise multiplication operation gives a single shared weight matrix additional expressive power, enabling feature modeling comparable to multiple independent matrices. It maintains nonlinear mapping capability while significantly reducing computational cost.

3) Simplified architecture reduces computational redundancy: LSTM relies on three gating structures for complex information filtering and memory management; TCN requires multiple layers of causal convolutions and residual connections; Transformer requires complex self-attention computations. In contrast, Memory-DD achieves equivalent memory management using two functionally specialized dendritic-inspired neuron groups, requiring only two matrix multiplications per time step (with shared weights), effectively avoiding computational redundancy across the architecture.

\section{Experiments and Analysis}
To systematically validate the effectiveness and computational efficiency of the proposed Memory-DD model, this section provides a comprehensive evaluation through multi-dimensional experimental analyses. The experiments are designed following strict control principles, directly comparing Memory-DD with mainstream time-series prediction models under a unified experimental environment to ensure objectivity and reliability of the results.

The organization of this section is as follows: First, we introduce the experimental setup, including the selection criteria for benchmark datasets, the technical characteristics of comparison models, the experimental environment configuration, and the hyperparameter settings to ensure reproducibility. Second, we clearly define the evaluation metrics used to quantify model prediction performance and computational complexity. Third, we present and analyze the results of computational complexity comparisons and detailed experimental results on time-series prediction tasks, including performance comparisons for both classification and regression tasks. Finally, we conduct ablation studies to verify the contributions of each core component of Memory-DD, providing empirical support for the rationale behind the model design.

\subsection{Experiment Settings}
\subsubsection{Comparison Models}
To evaluate the time-series prediction performance of the Memory-DD model, we selected a set of milestone or representative models in the field as performance benchmarks. These models cover the mainstream approaches for handling time-series data:Classical recurrent neural networks (RNNs): including the foundational Long Short-Term Memory network (LSTM) \cite{ref2} and its more computationally efficient variant, the Gated Recurrent Unit (GRU) \cite{ref21}.Enhanced RNN architectures: namely the Bidirectional LSTM (BiLSTM) \cite{ref26}, which can encode both past and future contextual information.Non-recurrent architectures: including the Time Convolutional Network (TCN) \cite{ref27}, which uses causal convolutions and performs well in sequence modeling, and the Transformer model \cite{ref3}, which completely abandons recurrent and convolutional structures and relies entirely on self-attention mechanisms.By directly comparing with these models, we can comprehensively assess the relative advantages of Memory-DD in terms of prediction accuracy and computational efficiency.

\subsubsection{Datasets}
To ensure both breadth and depth in evaluation, our experiments cover two core tasks: time-series classification and time-series regression. For this purpose, we carefully selected multiple publicly available benchmark datasets that are representative of the field.
Time-series classification datasets: We selected 18 diverse datasets from the UCR Time Series Classification Archive \cite{ref28}, which is widely recognized as the gold standard in the time-series data mining community. The chosen datasets span various real-world applications, including motion gesture recognition (Motion), electrocardiogram signal analysis (ECG), sensor readings (Sensor), and traffic flow monitoring (Traffic). The statistical information of these datasets is summarized in Table \Romannum{2}. They vary significantly in sequence length, data dimension, number of classes, and sample size, providing a comprehensive challenge for evaluating model generalization and robustness across different data characteristics. All training and test splits follow the official proportions provided by the UCR archive.

\begin{table*}[!t]
\centering
\caption{Time series Classification Dataset Statistics, The columns denote the size of the training set (Train) and testing set (Test), the number of classes (Class), the length of each time series (Length), the number of variables (Dimension), and the application domain (Type).}
\label{tab:dataset_stats}

\begin{tabular}{ccccccc}
\toprule
\textbf{Dataset} & \textbf{Train} & \textbf{Test} & \textbf{Class} & \textbf{Length} & \textbf{Dimension} & \textbf{Type} \\
\midrule
ArticularyWordRecognition [28] & 275 & 300 & 25 & 144 & 9 & Motion \\
BasicMotions [28] & 40 & 40 & 4 & 100 & 6 & HAR \\
Cricket [28] & 108 & 72 & 12 & 1197 & 6 & HAR \\
Ering [28] & 30 & 270 & 6 & 65 & 4 & HAR \\
Libras [28] & 180 & 180 & 15 & 45 & 2 & HAR \\
PenDigits [28] & 7494 & 3498 & 10 & 8 & 2 & Motion \\
ECG200 [28] & 100 & 100 & 2 & 96 & 1 & ECG \\
ECG5000 [28] & 500 & 4500 & 5 & 140 & 1 & ECG \\
BeetleFly [28] & 20 & 20 & 2 & 512 & 1 & Image \\
Chinatown [28] & 20 & 345 & 2 & 24 & 1 & Traffic \\
GunPointOldVersusYoung [28] & 135 & 316 & 2 & 150 & 1 & Motion \\
ItalyPowerDemand [28] & 67 & 1029 & 2 & 24 & 1 & Sensor \\
InsectEPGSmallTrain [28] & 17 & 249 & 3 & 601 & 1 & EPG \\
MoteStrain [28] & 20 & 1252 & 2 & 84 & 1 & Sensor \\
Plane [28] & 105 & 105 & 7 & 144 & 1 & Sensor \\
SonyAIBORobotSurface2 [28] & 27 & 953 & 2 & 65 & 1 & Sensor \\
ToeSegmentation2 [28] & 36 & 130 & 2 & 343 & 1 & Motion \\
SyntheticControl [28] & 300 & 300 & 6 & 60 & 1 & Simulated \\
\bottomrule
\end{tabular}
\end{table*}

Time-series regression datasets: We used nine publicly available benchmark datasets that are widely adopted in long-sequence time-series forecasting (LSTF) studies \cite{ref24,ref29,ref30,ref31}. Their detailed statistics are summarized in Table \Romannum{3}. These datasets were selected to ensure diversity and representativeness. The sources cover electricity consumption (Electricity, EETh1, EETh2, EETm1, EETm2), financial markets (Exchange-Rate), transportation (Traffic), meteorology (Weather), and epidemiology (illness), ensuring the evaluation has practical application value. The data sampling frequencies range from high-frequency (e.g., every 10 minutes) to low-frequency (e.g., weekly), requiring models to capture dynamic patterns at different time scales. The number of variables in the datasets ranges from a few (e.g., illness) to nearly a thousand (e.g., Traffic), testing the models’ ability to handle high-dimensional and complex time-series data. To strictly follow time-series forecasting protocols and prevent any leakage of future information, all regression datasets are split in chronological order, with the first 70\% of each sequence used as the training set and the remaining 30\% as the test set \cite{ref32}.

Based on this setup, to comprehensively evaluate model performance under different prediction horizons, we defined four prediction-step tasks for each dataset. Specifically, each task uses the past $L$ time steps ($L$ = 3, 6, 12, or 24) as input to predict the values of the next $P$ time steps ($P$ = 3, 6, 12, or 24) \cite{ref33}. During both training and testing, all sample constructions strictly follow the temporal order. Each sample consists of an input sequence of length $L$ and a target output sequence of length $P$. For example, in the task of predicting the next three time steps, each sample contains three time steps of input data and the subsequent three time steps as target labels.

\begin{table}[!t]
\centering
\caption{Time series regression Dataset Statistics. Length: length of time series; Dimension: number of variables; Sample-Rate: the sample rate.}
\label{tab:regression_dataset_stats}

\begin{tabular}{cccc}
\toprule
\textbf{Dataset} & \textbf{Length} & \textbf{Dimension} & \textbf{Sample-Rate} \\
\midrule
Electricity [29] & 26304 & 321 & 1 hour \\
Exchange\_Rate [29] & 7588 & 8 & 1 day \\
Illness [29] & 966 & 7 & 1 week \\
Traffic [31] & 17544 & 862 & 1 hour \\
Weather [30] & 52696 & 21 & 10 minutes \\
EETh1 [31] & 17420 & 7 & 1 hour \\
EETh2 [24] & 17420 & 7 & 1 hour \\
EETm1 [24] & 69680 & 7 & 15 minutes \\
EETm2 [24] & 69680 & 7 & 15 minutes \\
\bottomrule
\end{tabular}
\end{table}

All experiments were run on a computer with an Intel Core i9-14900HX CPU and an NVIDIA GeForce RTX 4060 Laptop GPU. We used the PyTorch 2.0.1+cu118 deep learning framework \cite{ref34}. To ensure a fair comparison, all models shared the same training settings. We used the Adam optimizer with a learning rate of 0.001 \cite{ref35}.For the time-series classification task, the batch size was set to 32. The hidden layer dimension was set to 128. For the regression task, the batch size was 128. The hidden layer dimension was 256.

\subsubsection{Evaluation Metrics}
To comprehensively assess model performance, we systematically evaluate all models from two orthogonal perspectives: prediction performance and computational complexity. We use a set of standardized quantitative metrics that focus respectively on predictive effectiveness and computational efficiency. For different task types, we select appropriate standard metrics to evaluate model prediction performance.

For time-series classification tasks, we use accuracy \cite{ref36} and F1 score \cite{ref37} as performance evaluation metrics. Accuracy is the most intuitive metric in classification, representing the proportion of correctly predicted samples. The F1 score \cite{ref37} is the harmonic mean of precision \cite{ref37} and recall \cite{ref37}, making it particularly suitable for datasets with imbalanced class distributions. It serves as a key complementary metric for evaluating classification effectiveness. The specific definitions are as follows:
\begin{equation}
Accuracy = \frac{TP + TN}{TP + TN + FP + FN}
\end{equation}

Here, $TP$ denotes the number of samples that are actually positive and correctly predicted as positive; $TN$ denotes the number of samples that are actually negative and correctly predicted as negative; $FP$ denotes the number of samples that are actually negative but incorrectly predicted as positive; $FN$ denotes the number of samples that are actually positive but incorrectly predicted as negative.
\begin{equation}
F1-Score = 2 \times \frac{Precision \times Recall}{Precision + Recall}
\end{equation}

Here, $Precision$ represents the proportion of samples predicted as positive that are actually positive. $Recall$ represents the proportion of actual positive samples that are correctly predicted as positive. The formulas are given as follows:
\begin{equation}
Precision = \frac{TP}{TP + FP}
\end{equation}

\begin{equation}
Recall = \frac{TP}{TP + FN}
\end{equation}

For time-series regression tasks, Mean Squared Error (MSE) \cite{ref38} is used as the performance metric. It calculates the average of the squared differences between the predicted and true values. This metric is sensitive to outliers (i.e., large prediction errors) and can effectively reflect the stability of the model’s predictions \cite{ref39}. The formula is as follows:
\begin{equation}
MSE = \frac{1}{N} \sum_{i=1}^{N} (y_i - \hat{y}_i)^2
\end{equation}

Here, $N$ denotes the total number of samples in the test set, $y_i$ represents the true value of the $i$-th sample, and $\hat{y}_i$ represents the predicted value of the $i$-th sample.

In terms of computational complexity, we use the total number of parameters and floating-point operations (FLOPs) as evaluation metrics \cite{ref40}. The total number of parameters measures the overall learnable parameters of the model, which directly determines the model size and storage requirements. FLOPs quantify the total floating-point operations required for a single forward pass, serving as a theoretical standard for measuring computational cost and generally correlating positively with the model’s inference speed \cite{ref41}. The formulas are as follows:
\begin{equation}
Params = \sum_{l=1}^{L} (W_l + b_l)
\end{equation}

Here, $L$ denotes the total number of layers in the model, $W_l$ represents the number of weight parameters in the l-th layer, and $b_l$ represents the number of bias parameters in the l-th layer.

For floating-point operations, taking a fully connected layer as an example, the FLOPs of the l fully connected layer are calculated as follows:
\begin{equation}
FLOPs_l = 2 \times d_{in} \times d_{out} -d_{out}
\end{equation}

Here, $d_{in}$ denotes the input dimension of the fully connected layer, and $d_{out}$ denotes the output dimension.

The total FLOPs of the model are:
\begin{equation}
FLOPs = \sum_{l=1}^{L} FLOPs_l
\end{equation}

An ideal efficient model should achieve competitive predictive performance while exhibiting lower total parameters and FLOPs.

\begin{figure*}[!t]
\centering
\includegraphics[width=5.8in]{Fig2.pdf}
\caption{Comparison of model complexity for the baseline models and Memory-DD (Ours). (a) The number of parameters (in millions). (b) The computational cost in Giga Floating Point Operations (G-FLOPs).}
\label{fig_2}
\end{figure*}

\subsection{Computational Complexity Experimental Results}
To validate the lightweight design of the Memory-DD model, we used PyTorch’s built-in toolkit to calculate its total number of parameters and FLOPs.Table \Romannum{4} and Fig. 2 provide a detailed comparison of Memory-DD and baseline models in terms of total parameters (millions, M) and floating-point operations (gigaFLOPs, G). The results show that the parameter count of Memory-DD is approximately 65.5\% of GRU (0.055 M) and 50\% of LSTM (0.072 M), and is much lower than BiLSTM (0.145 M), TCN (0.353 M), and Transformer (0.597 M). This extremely low parameter count provides significant advantages in storage, deployment, and overfitting prevention, making it particularly suitable for resource-constrained edge devices. In terms of FLOPs, Memory-DD (0.0073 G) is even slightly lower than GRU (0.0076 G) by about 3.9\%, and reduces approximately 27.7\% compared to LSTM (0.0101 G). Compared with TCN (0.0537 G) and Transformer (0.0855 G), its FLOPs are reduced by about 86.4\% and 91.5\%, respectively. This directly translates into faster inference speed and lower energy consumption, which is critical for time-sensitive sequence prediction applications.

\begin{table}[!t]
\centering
\caption{Statistics of Parameters (in Millions, M) and FLOPs (in Giga, G) for different models.}
\label{tab:model_stats}

\begin{tabular}{ccc}
\toprule
\textbf{Models} & \textbf{Parameters (M)} & \textbf{FLOPs (G)} \\
\midrule
LSTM & 0.072 & 0.0101 \\
GRU & 0.055 & 0.0076 \\
BiLSTM & 0.145 & 0.0202 \\
TCN & 0.353 & 0.0537 \\
Transformer & 0.597 & 0.0855 \\
\textbf{Memory-DD (Ours)} & \textbf{0.036} & \textbf{0.0073} \\
\bottomrule
\end{tabular}
\end{table}

The reason Memory-DD achieves such low computational complexity lies in its innovative dendritic-inspired neuron group design. Traditional recurrent neural networks, such as LSTM and GRU, rely on matrix multiplications and nonlinear activation functions like Sigmoid and Tanh to introduce complex feature interactions and gating mechanisms. These operations result in a large number of parameters and FLOPs.Memory-DD, inspired by the multiplicative gating mechanism of biological dendrites, uses the Hadamard product—an element-wise multiplication—to realize complex nonlinear mappings. Mathematically, the Hadamard product requires far fewer parameters and FLOPs than fully connected layers, while still providing strong nonlinear computational capabilities without introducing traditional complex nonlinear functions. A Memory-DD unit consists of two streamlined dendritic-inspired neuron groups, avoiding the multiple gates (input, forget, output) and redundant matrix operations found in LSTM and GRU. This fundamentally reduces computational overhead at the structural level. Such a design allows Memory-DD to achieve high representational capacity while maintaining minimal computational cost.

\subsection{Time-Series Prediction Experimental Results}
This section presents the detailed experimental results of Memory-DD on time-series classification and regression tasks, comparing it with baseline models and discussing the findings.

\subsubsection{Classification Task Results}
We systematically evaluated the prediction performance of Memory-DD and comparison models on 18 UCR time-series classification datasets. The results are shown in Table \Romannum{5}. Statistical analysis indicates that Memory-DD demonstrates significant advantages in overall performance metrics, achieving an average accuracy of 89.41\% and an average F1 score of 85.12\%.

\begin{table*}[t]
\centering
\caption{Performance evaluation of the models on 18 UCR time-series classification datasets, using Accuracy (abbreviated as ``Acc'') and F1-Score (abbreviated as ``F1'') as metrics.}
\label{tab:performance_comparison}

\begin{tabular}{
    l
    *{12}{S[table-format=1.4]}
}
\toprule
 & \multicolumn{2}{c}{LSTM} & \multicolumn{2}{c}{GRU} & \multicolumn{2}{c}{BiLSTM} & \multicolumn{2}{c}{TCN} & \multicolumn{2}{c}{Transformer} & \multicolumn{2}{c}{Memory-DD (Ours)} \\
\cmidrule(lr){2-3} \cmidrule(lr){4-5} \cmidrule(lr){6-7} \cmidrule(lr){8-9} \cmidrule(lr){10-11} \cmidrule(lr){12-13}
Dataset & {Acc} & {F1} & {Acc} & {F1} & {Acc} & {F1} & {Acc} & {F1} & {Acc} & {F1} & {Acc} & {F1} \\
\midrule
ArticularyWordRecognition & 0.9133 & 0.9091 & {\bfseries 0.9167} & {\bfseries 0.9139} & 0.9133 & 0.9094 & {\bfseries 0.9367} & {\bfseries 0.9330} & {\bfseries 0.9800} & {\bfseries 0.9805} & 0.8633 & 0.8607 \\
BasicMotions & 0.7000 & 0.6958 & 0.7500 & 0.7170 & 0.6750 & 0.6613 & {\bfseries 0.9500} & {\bfseries 0.9499} & {\bfseries 0.9750} & {\bfseries 0.9749} & {\bfseries 0.9000} & {\bfseries 0.8990} \\
Cricket & 0.6389 & 0.5555 & 0.7917 & 0.7708 & 0.7361 & 0.7191 & {\bfseries 0.9583} & {\bfseries 0.9581} & {\bfseries 0.9167} & {\bfseries 0.9143} & {\bfseries 0.9537} & {\bfseries 0.9541} \\
ERing & 0.7556 & 0.7528 & 0.7481 & 0.7377 & 0.7556 & 0.7298 & {\bfseries 0.9185} & {\bfseries 0.9183} & {\bfseries 0.9444} & {\bfseries 0.9443} & {\bfseries 0.8000} & {\bfseries 0.7946} \\
Libras & 0.7722 & 0.7719 & {\bfseries 0.8333} & {\bfseries 0.8326} & 0.7722 & 0.7670 & {\bfseries 0.8389} & {\bfseries 0.8353} & {\bfseries 0.9056} & {\bfseries 0.9024} & 0.7556 & 0.7516 \\
PenDigits & {\bfseries 0.9860} & {\bfseries 0.9860} & 0.9851 & 0.9852 & 0.9854 & 0.9854 & {\bfseries 0.9880} & {\bfseries 0.9880} & 0.9794 & 0.9793 & {\bfseries 0.9857} & {\bfseries 0.9857} \\
ECG200 & 0.8000 & 0.7702 & 0.7900 & 0.7643 & 0.8100 & 0.7835 & {\bfseries 0.8800} & {\bfseries 0.8681} & {\bfseries 0.8700} & {\bfseries 0.8581} & {\bfseries 0.8200} & {\bfseries 0.7932} \\
ECG5000 & 0.9378 & 0.5567 & 0.9358 & 0.5509 & 0.9380 & {\bfseries 0.5568} & {\bfseries 0.9413} & {\bfseries 0.5807} & {\bfseries 0.9407} & {\bfseries 0.5952} & {\bfseries 0.9384} & 0.5494 \\
BeetleFly & {\bfseries 0.8500} & {\bfseries 0.8496} & 0.8000 & 0.7917 & {\bfseries 0.8500} & {\bfseries 0.8496} & 0.7500 & 0.7333 & {\bfseries 0.9000} & {\bfseries 0.9000} & 0.7500 & 0.7333 \\
Chinatown & 0.8309 & 0.7941 & 0.9767 & 0.9709 & 0.8309 & 0.7916 & {\bfseries 0.9825} & {\bfseries 0.9784} & {\bfseries 0.9854} & {\bfseries 0.9817} & {\bfseries 0.9825} & {\bfseries 0.9780} \\
GunPointOldVersusYoung & 1.0000 & 1.0000 & 1.0000 & 1.0000 & 1.0000 & 1.0000 & 1.0000 & 1.0000 & 1.0000 & 1.0000 & 0.9587 & 0.9584 \\
ItalyPowerDemand & 0.9699 & 0.9699 & {\bfseries 0.9708} & {\bfseries 0.9708} & 0.9689 & 0.9689 & {\bfseries 0.9738} & {\bfseries 0.9738} & 0.9611 & 0.9611 & {\bfseries 0.9738} & {\bfseries 0.9738} \\
InsectEPGSmallTrain & {\bfseries 1.0000} & {\bfseries 1.0000} & {\bfseries 1.0000} & {\bfseries 1.0000} & {\bfseries 1.0000} & {\bfseries 1.0000} & {\bfseries 1.0000} & {\bfseries 1.0000} & {\bfseries 1.0000} & {\bfseries 1.0000} & {\bfseries 1.0000} & {\bfseries 1.0000} \\
MoteStrain & 0.8115 & 0.8058 & {\bfseries 0.8203} & {\bfseries 0.8177} & {\bfseries 0.8307} & {\bfseries 0.8267} & 0.7684 & 0.7597 & {\bfseries 0.8898} & {\bfseries 0.8889} & 0.7915 & 0.7915 \\
Plane & 0.7333 & 0.7211 & 0.6190 & 0.5824 & 0.8286 & 0.8264 & {\bfseries 0.9905} & {\bfseries 0.9897} & {\bfseries 0.9905} & {\bfseries 0.9908} & {\bfseries 0.9810} & {\bfseries 0.9740} \\
SonyAIBORobotSurface2 & 0.8248 & 0.8133 & 0.7597 & 0.7370 & 0.8195 & 0.8083 & {\bfseries 0.8909} & {\bfseries 0.8833} & {\bfseries 0.8374} & {\bfseries 0.8295} & {\bfseries 0.8321} & {\bfseries 0.8176} \\
ToeSegmentation2 & 0.7846 & 0.6024 & 0.8077 & {\bfseries 0.6754} & 0.7615 & {\bfseries 0.6327} & {\bfseries 0.8462} & {\bfseries 0.6436} & {\bfseries 0.8308} & 0.5299 & {\bfseries 0.8308} & 0.5299 \\
SyntheticControl & {\bfseries 0.9933} & {\bfseries 0.9933} & 0.9933 & 0.9933 & 0.9933 & 0.9933 & {\bfseries 1.0000} & {\bfseries 1.0000} & {\bfseries 0.9967} & {\bfseries 0.9967} & 0.9767 & 0.9766 \\
\midrule
Average & 0.8516 & 0.8064 & 0.8632 & 0.8248 & 0.8609 & 0.8232 & {\bfseries 0.9204} & {\bfseries 0.8838} & {\bfseries 0.9356} & {\bfseries 0.8962} & {\bfseries 0.8941} & {\bfseries 0.8612} \\
\midrule 
Avg Acc / Param (M) & {\bfseries 11.83} & {{--}} & {\bfseries 15.69} & {{--}} & 5.93 & {{--}} & 2.61 & {{--}} & 1.57 & {{--}} & {\bfseries \textcolor{red}{24.53}} & {{--}} \\
Avg Acc / FLOPs (G) & {\bfseries 84.32} & {{--}} & {\bfseries 113.58} & {{--}} & 42.6 & {{--}} & 17.14 & {{--}} & 10.9 & {{--}} & {\bfseries \textcolor{red}{122.48}} & {{--}} \\
\bottomrule
\end{tabular}

\smallskip 
\footnotesize \textbf{Table Notes:} \raggedright For each dataset, the results of the top three performing models are highlighted in bold. ``Avg acc / Param (M)'' quantifies the parameter efficiency, indicating the accuracy improvement per million parameters. ``Avg acc / FLOPs (G)'' quantifies the computational efficiency, indicating the accuracy improvement per Giga FLOPs. TCN and Transformer exhibit \textbf{extremely high model complexity} (9.81$\times$ and 16.58$\times$ the parameters of Memory-DD, respectively).
\end{table*}

Compared with traditional recurrent neural network architectures, Memory-DD achieves performance improvements on most datasets. Specifically, compared with LSTM (average accuracy 85.16\%) and GRU (average accuracy 86.32\%), Memory-DD achieves improvements of 4.25\% and 2.59\%, respectively. This result suggests that the dendritic-inspired neuron design of Memory-DD can more effectively capture nonlinear temporal dependencies in time-series data.

\begin{figure*}[!t]
\centering  
\includegraphics[width=0.9\textwidth]{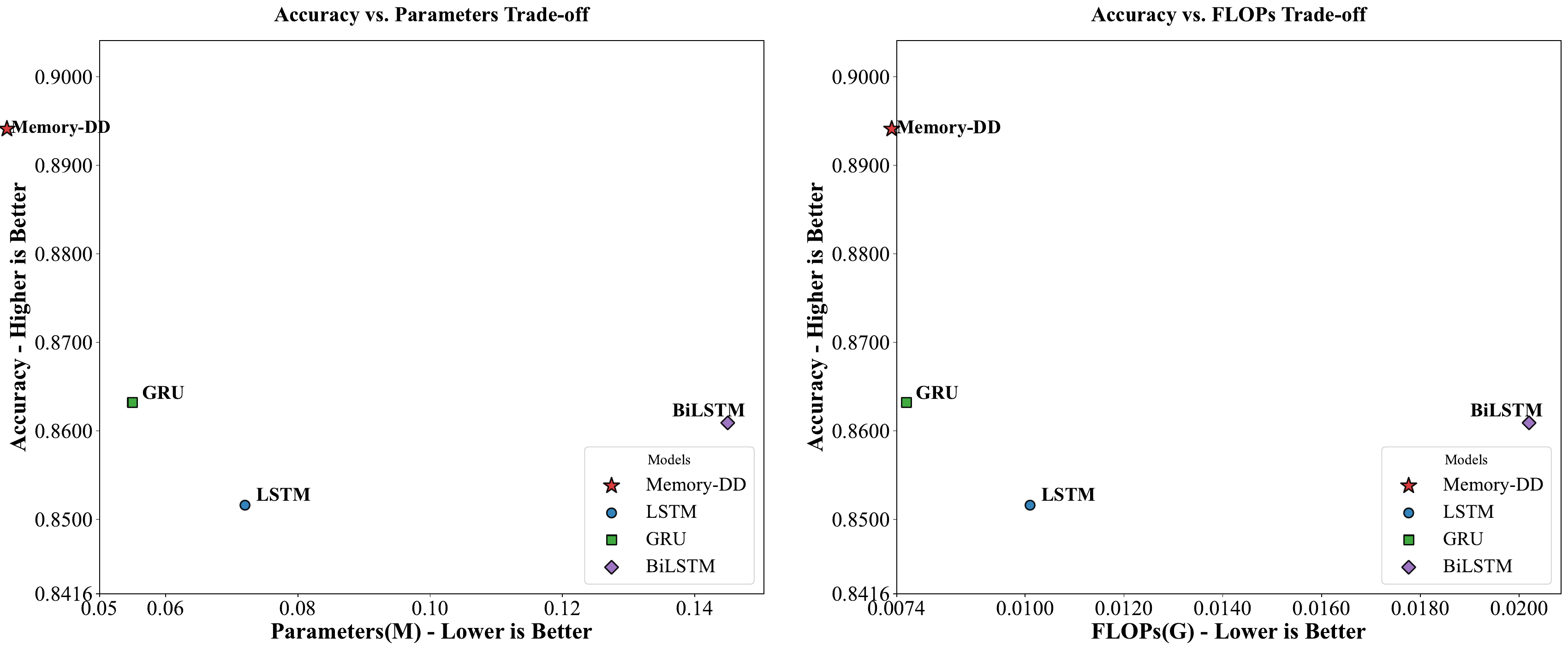}
\caption{The trade-off between average accuracy and model complexity. (a) Left figure: Average accuracy plotted against the number of parameters (M). (b) Right figure: Average accuracy plotted against the computational cost in Giga Floating Point Operations (G-FLOPs). The top-left corner represents the ideal balance of high accuracy and low complexity.}
\label{fig_6}
\end{figure*}

In comparisons with more complex architectures, although TCN (92.04\%) and Transformer (93.56\%) slightly outperform Memory-DD in average accuracy, considering the large differences in computational complexity, Memory-DD offers a superior performance-efficiency trade-off. Notably, on the Cricket dataset, Memory-DD reaches 95.37\% accuracy, representing a substantial improvement over LSTM’s 63.89\%. This may be attributed to the complex multivariate temporal interactions present in this dataset, which are well captured by Memory-DD’s Hadamard product gating mechanism.

To quantify model efficiency, we introduce a parameter efficiency metric (accuracy/parameter count) and a computational efficiency metric (accuracy/FLOPs). Memory-DD achieves a parameter efficiency of 24.53 $M^{-1}$, significantly outperforming LSTM (11.83 $M^{-1}$), GRU (15.69 $M^{-1}$), TCN (2.61 $M^{-1}$), and Transformer (1.57 $M^{-1}$). In terms of computational efficiency, Memory-DD achieves a value of 122.48 $G^{-1}$, ranking first as well.

To more clearly illustrate the performance-complexity trade-off, Fig. 3 shows a scatter plot of different models’ accuracy versus computational complexity. It should be noted that due to the much higher computational complexity of TCN (0.353M parameters, 0.0537G FLOPs) and Transformer (0.597M parameters, 0.0855G FLOPs), these two models are excluded from Fig. 3 to ensure effective comparison, focusing on Memory-DD versus traditional recurrent neural networks. From Fig. 3(a), which analyzes the parameter-accuracy trade-off, Memory-DD is located in the upper-left region, achieving the optimal combination of low parameter count (0.036M) and high accuracy (89.41\%). In contrast, LSTM, GRU, and BiLSTM achieve similar accuracy but with substantially higher parameter counts. Fig. 3(b), showing FLOPs versus accuracy, further confirms this advantage: Memory-DD maintains competitive prediction accuracy while incurring the lowest computational cost.

This significant efficiency advantage stems from Memory-DD’s core design principle: replacing traditional matrix multiplications with Hadamard products, which greatly reduces computational complexity. Traditional LSTM relies on multiple gated units (forget gate, input gate, output gate) for complex matrix operations, whereas Memory-DD achieves comparable memory update and decision-making functionality using only two dendritic-inspired neuron groups.

Further analysis of performance across datasets reveals that Memory-DD performs exceptionally well on datasets with clear temporal patterns (e.g., BasicMotions, Cricket), while performance is relatively moderate on datasets with higher noise or simpler patterns (e.g., BeetleFly). This phenomenon may be related to Memory-DD’s nonlinear feature interaction mechanism: when complex feature interactions exist in the time-series data, Hadamard product operations can effectively capture these relationships; conversely, for linearly separable or noise-dominated data, traditional methods may be more straightforwardly effective.

\subsubsection{Regression Task Results}
The performance of Memory-DD on nine time-series regression datasets across different prediction horizons (3, 6, 12, 24 steps) is shown in Table \Romannum{6}. Overall, Memory-DD achieves an average MSE of 0.0046, which is only 0.0005 higher than the best baseline model, BiLSTM (0.0041), while its parameter count is merely 24.8\% of BiLSTM’s.

In terms of absolute prediction accuracy, Memory-DD maintains comparable performance to traditional models on most datasets. On the Electricity dataset, Memory-DD achieves MSEs ranging from 0.0033 to 0.0046 across all prediction horizons, essentially matching LSTM (0.0033–0.0045) and GRU (0.0033–0.0046). On the ETTm2 dataset, Memory-DD also demonstrates stable prediction capability, with MSEs ranging from 0.0005 to 0.0018.

\newcolumntype{C}{>{\centering\arraybackslash}X}

\begin{table*}[htbp]
    \centering
    \caption{Comparison of Mean Squared Error (MSE) results for each model on 9 time-series regression datasets across different prediction horizons (3, 6, 12, 24), The top three models by average accuracy are shown in bold.}
    \label{tab:comparison}
    \begin{tabularx}{\textwidth}{l c *{6}{C}}
        \toprule
        \multirow{2}{*}{Dataset} & \multirow{2}{*}{Horizon} & \multicolumn{6}{c}{Metrics (MSE)} \\
        \cmidrule(lr){3-8}
        & & LSTM & GRU & BiLSTM & TCN & Transformer & \makebox[0pt]{Memory-DD (Ours)} \\
        \midrule
        
        \multirow{4}{*}{Electricity}
        & 3  & 0.0033 & 0.0033 & 0.0031 & 0.0064 & 0.0038 & 0.0033 \\
        & 6  & 0.0039 & 0.0038 & 0.0037 & 0.0054 & 0.0042 & 0.0038 \\
        & 12 & 0.0044 & 0.0043 & 0.0044 & 0.0054 & 0.0045 & 0.0044 \\
        & 24 & 0.0045 & 0.0046 & 0.0031 & 0.0064 & 0.0047 & 0.0046 \\
        \midrule
        
        \multirow{4}{*}{Exchange-Rate}
        & 3  & 0.0005 & 0.0003 & 0.0003 & 0.0259 & 0.0077 & 0.0004 \\
        & 6  & 0.0010 & 0.0006 & 0.0008 & 0.0200 & 0.0066 & 0.0014 \\
        & 12 & 0.0027 & 0.0010 & 0.0019 & 0.0184 & 0.0069 & 0.0040 \\
        & 24 & 0.0061 & 0.0017 & 0.0032 & 0.0200 & 0.0094 & 0.0063 \\
        \midrule 
        
        \multirow{4}{*}{Illness}
        & 3  & 0.0059 & 0.0203 & 0.0056 & 0.0237 & 0.0106 & 0.0063 \\
        & 6  & 0.0111 & 0.0158 & 0.0101 & 0.0273 & 0.0133 & 0.0101 \\
        & 12 & 0.0195 & 0.0097 & 0.0166 & 0.0320 & 0.0181 & 0.0166 \\
        & 24 & 0.0238 & 0.0053 & 0.0219 & 0.0431 & 0.0245 & 0.0201 \\
        \midrule
        
        \multirow{4}{*}{Traffic}
        & 3  & 0.0035 & 0.0036 & 0.0035 & 0.0050 & 0.0034 & 0.0036 \\
        & 6  & 0.0040 & 0.0041 & 0.0040 & 0.0054 & 0.0039 & 0.0041 \\
        & 12 & 0.0039 & 0.0040 & 0.0038 & 0.0146 & 0.0037 & 0.0040 \\
        & 24 & 0.0039 & 0.0040 & 0.0039 & 0.0146 & 0.0038 & 0.0041 \\
        \midrule
        
        \multirow{4}{*}{Weather}
        & 3  & 0.0019 & 0.0019 & 0.0019 & 0.0044 & 0.0020 & 0.0019 \\
        & 6  & 0.0023 & 0.0023 & 0.0023 & 0.0039 & 0.0024 & 0.0023 \\
        & 12 & 0.0030 & 0.0029 & 0.0030 & 0.0040 & 0.0030 & 0.0030 \\
        & 24 & 0.0039 & 0.0039 & 0.0040 & 0.0051 & 0.0043 & 0.0041 \\
        \midrule
        
        \multirow{4}{*}{ETTh1}
        & 3  & 0.0056 & 0.0055 & 0.0055 & 0.0099 & 0.0053 & 0.0056 \\
        & 6  & 0.0078 & 0.0074 & 0.0080 & 0.0113 & 0.0073 & 0.0078 \\
        & 12 & 0.0085 & 0.0085 & 0.0044 & 0.0100 & 0.0085 & 0.0090 \\
        & 24 & 0.0080 & 0.0081 & 0.0031 & 0.0089 & 0.0083 & 0.0080 \\
        \midrule
        
        \multirow{4}{*}{ETTh2}
        & 3  & 0.0013 & 0.0013 & 0.0013 & 0.0046 & 0.0016 & 0.0013 \\
        & 6  & 0.0018 & 0.0017 & 0.0018 & 0.0041 & 0.0021 & 0.0018 \\
        & 12 & 0.0022 & 0.0022 & 0.0023 & 0.0035 & 0.0024 & 0.0024 \\
        & 24 & 0.0026 & 0.0025 & 0.0026 & 0.0043 & 0.0029 & 0.0029 \\
        \midrule
        
        \multirow{4}{*}{ETTm1}
        & 3  & 0.0017 & 0.0017 & 0.0017 & 0.0075 & 0.0018 & 0.0017 \\
        & 6  & 0.0025 & 0.0026 & 0.0027 & 0.0059 & 0.0025 & 0.0027 \\
        & 12 & 0.0042 & 0.0040 & 0.0040 & 0.0052 & 0.0037 & 0.0041 \\
        & 24 & 0.0065 & 0.0065 & 0.0069 & 0.0072 & 0.0061 & 0.0070 \\
        \midrule
        
        \multirow{4}{*}{ETTm2}
        & 3  & 0.0005 & 0.0005 & 0.0005 & 0.0037 & 0.0007 & 0.0005 \\
        & 6  & 0.0007 & 0.0007 & 0.0007 & 0.0029 & 0.0008 & 0.0007 \\
        & 12 & 0.0010 & 0.0010 & 0.0010 & 0.0021 & 0.0011 & 0.0011 \\
        & 24 & 0.0017 & 0.0015 & 0.0016 & 0.0026 & 0.0018 & 0.0018 \\
        \midrule
        
        \multicolumn{2}{l}{Average} 
        & 0.0047 & \bfseries 0.0043 & \bfseries 0.0041 & 0.0107 & 0.0055 & \bfseries 0.0046 \\
        \midrule 
        
        \multicolumn{8}{l}{\textbf{The number of parameters of the model / the number of parameters of Memory-DD}} \\
        \multicolumn{2}{l}{LSTM}       & & & & & 200.00\% \\
        \multicolumn{2}{l}{GRU}        & & & & & 152.78\% \\
        \multicolumn{2}{l}{BiLSTM}     & & & & & 402.78\% \\
        \multicolumn{2}{l}{TCN}        & & & & & 980.56\% \\
        \multicolumn{2}{l}{Transformer} & & & & & 1658.33\% \\
        \multicolumn{2}{l}{Memory-DD}  & & & & & 1 \\
        \midrule
        
        \multicolumn{8}{l}{\textbf{The FLOPs of the model / the FLOPs of Memory-DD}} \\
        \multicolumn{2}{l}{LSTM}       & & & & & 138.36\% \\
        \multicolumn{2}{l}{GRU}        & & & & & 104.11\% \\
        \multicolumn{2}{l}{BiLSTM}     & & & & & 276.71\% \\
        \multicolumn{2}{l}{TCN}        & & & & & 735.62\% \\
        \multicolumn{2}{l}{Transformer} & & & & & 1171.23\% \\
        \multicolumn{2}{l}{Memory-DD}  & & & & & 1 \\
        \bottomrule
    \end{tabularx}
\end{table*}

\begin{figure*}[!t]
\centering
\includegraphics[width=5.8in]{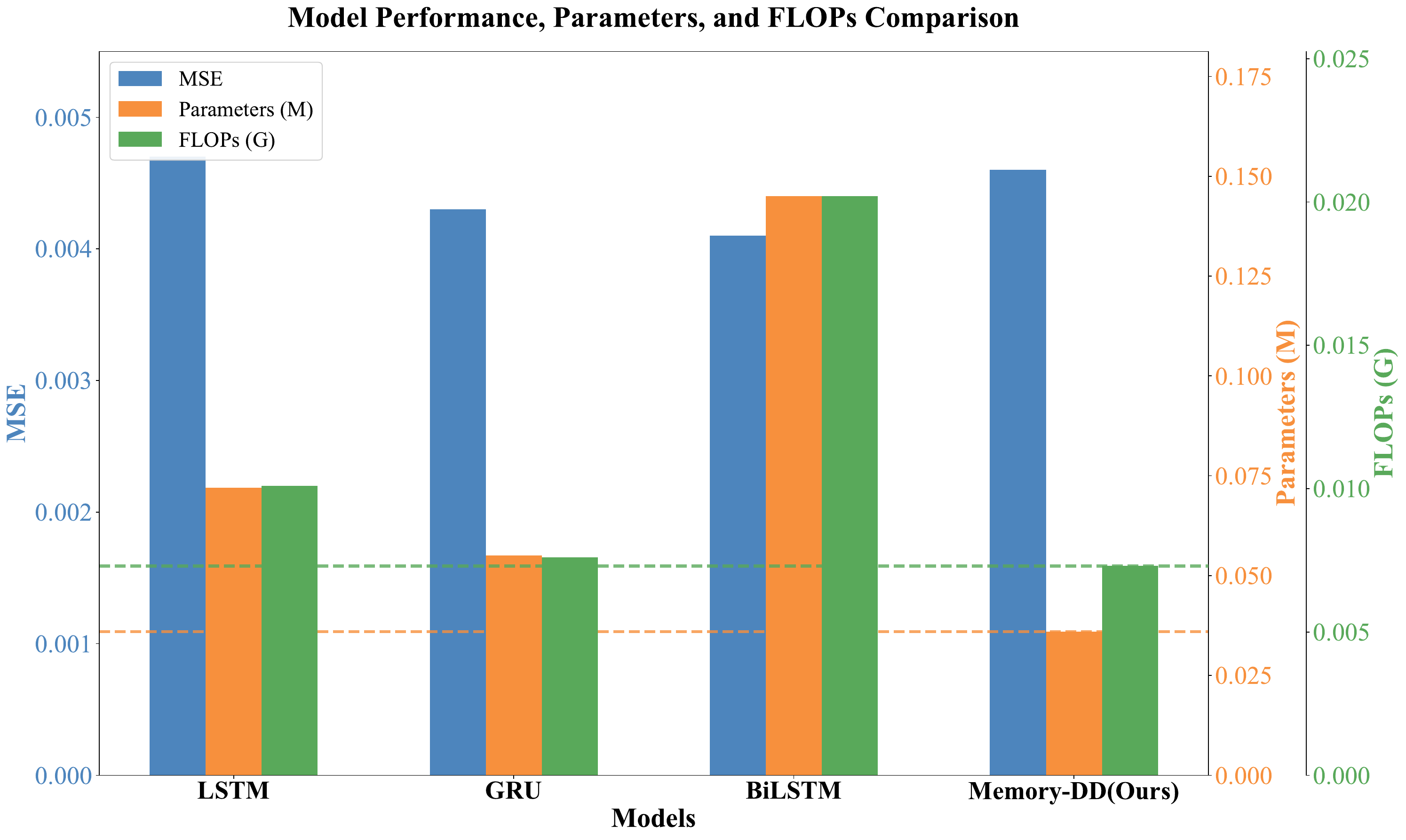}
\caption{Model Performance, Parameters, and FLOPS Comparison for LSTM, GRU, BiLSTM, and Memory-DD models.}
\label{fig_5}
\end{figure*}

However, on some datasets such as Exchange-Rate, Memory-DD exhibits a higher MSE of 0.0063 for long-term prediction (24 steps), compared to GRU’s 0.0017. This discrepancy may indicate limitations of Memory-DD in modeling non-stationarity and high-frequency noise in financial time-series data. The random walk characteristics and abrupt fluctuations typical of financial data may exceed the current memory mechanism’s modeling capacity.

To more intuitively illustrate the computational efficiency advantage of Memory-DD in regression tasks, the bottom of Table \Romannum{6} shows the relative computational complexity of each model compared to Memory-DD. The data indicate that LSTM has twice the parameters of Memory-DD (200\%) and 1.38 times the FLOPs (138.36\%); GRU has 1.53 times the parameters (152.78\%) and 1.04 times the FLOPs (104.11\%); BiLSTM has 4.03 times the parameters (402.78\%) and 2.77 times the FLOPs (276.71\%). These quantitative comparisons clearly demonstrate that Memory-DD achieves a significant reduction in computational complexity while maintaining comparable predictive performance. Notably, compared to BiLSTM, Memory-DD reduces parameter count and FLOPs by 75.2\% and 63.9\%, respectively, highlighting its practical value in resource-constrained deployment scenarios.

Fig. 4 further visualizes the comprehensive comparison of LSTM, GRU, BiLSTM, and Memory-DD across three dimensions: average MSE, parameter count, and FLOPs. From the figure, it is observed that Memory-DD (blue bars) achieves MSE performance comparable to traditional models, with an average MSE of 0.0046, slightly higher than the best-performing BiLSTM (0.0041) and GRU (0.0043). However, in terms of parameter count (orange bars) and FLOPs (green bars), Memory-DD shows a clear advantage, with values significantly lower than the other three models.

Results across different prediction horizons reveal the time-scale characteristics of Memory-DD. Statistical analysis shows that, as the prediction horizon increases, the MSE of most models rises, which aligns with general trends in time-series forecasting. Memory-DD maintains relatively stable performance for short-term (3–6 steps) and mid-term (12 steps) predictions, but its performance declines more noticeably for long-term predictions (24 steps). For example, on the Traffic dataset, Memory-DD’s MSE increases from 0.0036 for 3-step predictions to 0.0041 for 24-step predictions, slightly higher than GRU and BiLSTM. This time-scale dependence in predictive capability may be related to Memory-DD’s fixed memory update mechanism.

\subsection{Ablation Study}

\newcommand{\redtimes}{\textcolor{red}{\texttimes}}
\begin{table*}[htbp]
    \centering
    \caption{Ablation study results for different components of Memory-DD on the ItalyPowerDemand and Cricket datasets.}
    \label{tab:ablation_study}
    \begin{tabular}{l c c c c c c c c}
        \toprule
        \multirow{2}{*}{Models} & \multirow{2}{*}{Hadmard product} & \multicolumn{2}{c}{Residual Connection} & \multirow{2}{*}{Weight sharing} & \multicolumn{2}{c}{ItalyPowerDemand} & \multicolumn{2}{c}{Cricket} \\
        \cmidrule(lr){3-4} \cmidrule(lr){6-7} \cmidrule(lr){8-9}
        & & $C_t$ & $H_t$ & & Accuracy & F1-Score & Accuracy & F1-Score \\
        \midrule
        Memory-DD(Baseline) & \checkmark & \checkmark & \checkmark & \checkmark & 0.9738 & 0.9738 & 0.9722 & 0.9720 \\
        Variant A & \redtimes & \checkmark & \checkmark & \checkmark & 0.5656 & 0.4684 & 0.2500 & 0.1386 \\
        Variant B & \checkmark & \redtimes & \checkmark & \checkmark & 0.5015 & 0.3340 & 0.0833 & 0.0128 \\
        Variant C & \checkmark & \checkmark & \redtimes & \checkmark & 0.9689 & 0.9689 & 0.6389 & 0.5921 \\
        Variant D & \checkmark & \redtimes & \redtimes & \checkmark & 0.5015 & 0.3340 & 0.0833 & 0.0128 \\
        Variant E & \checkmark & \checkmark & \checkmark & \redtimes & 0.9738 & 0.9738 & 0.9583 & 0.9580 \\
        \bottomrule
    \end{tabular}
\end{table*}
To validate the contribution of each component in Memory-DD to overall performance, ablation experiments were conducted on two representative datasets, ItalyPowerDemand and Cricket, with results shown in Table \Romannum{7}. Five variant models were designed, each removing or disabling key components of Memory-DD. The ablation results reveal the critical roles of each component in the Memory-DD architecture. The Hadamard product operation is confirmed as the most essential component. Variant A (removing the Hadamard product) shows severe performance degradation on both datasets, with accuracy dropping from the baseline 97.38\% and 97.22\% to 56.56\% and 25.00\%, respectively. This indicates that the non-linear feature interaction mechanism of the dendrite-like units is fundamental to the model’s effectiveness.The importance of the residual connection mechanism is also validated. Variants B, C, and D (removing the residual connections of dendrite-like neuron group 1, group 2, and all residual connections, respectively) demonstrate that residual connections play an irreplaceable role in gradient propagation and information retention. In particular, Variant B (removing the residual connection of dendrite-like neuron group 2) and Variant D (removing all residual connections) both reduce accuracy to 8.33\% on the Cricket dataset, highlighting the importance of direct transmission of current information for final decision-making.Ablation of the weight-sharing strategy confirms the effectiveness of parameter compression. Variant E (removing weight sharing) maintains performance comparable to the baseline (97.38\%) on ItalyPowerDemand and shows a slight decrease on Cricket (from 97.22\% to 95.83\%). This indicates that the weight-sharing strategy significantly reduces parameter count with minimal loss in performance. Without weight sharing, the two dendrite-like neuron groups have independent weight matrices, increasing the total parameters. Therefore, the effectiveness of weight sharing lies in achieving substantial parameter efficiency with minimal impact on predictive performance.

\section{Discussion}
The proposed Memory-DD model represents an important extension of dendrite-inspired neurons from static data processing to dynamic temporal modeling. This study systematically validates the effectiveness and practical value of the biologically inspired computational paradigm in time-series prediction tasks. The core innovation of Memory-DD lies in organically combining the multiplicative gating mechanism of biological neuron dendrites with the memory mechanism of recurrent neural networks. This design overcomes the limitations of traditional temporal models that rely on complex activation functions and multiple gating structures. By using the Hadamard product—a simple yet powerful operation—Memory-DD achieves efficient nonlinear feature interactions. From a computational neuroscience perspective, this approach is closer to the actual working mechanism of biological neural systems, providing new ideas for building more biologically plausible artificial neural networks.

Comprehensive experimental results show that Memory-DD achieves comparable predictive performance to traditional models while significantly reducing parameter count and computational complexity. This outcome not only validates the effectiveness of biologically inspired architectural design but also opens a novel path for lightweight temporal modeling. In the context of deep learning models facing excessive computational costs and energy consumption, Memory-DD offers a feasible solution.

\section{Conclusion}
This study proposes the Memory-DD, a low-complexity dendrite-inspired neuron. By innovatively integrating the Hadamard product gating mechanism of biological dendrites with recurrent memory functionality, the model can extract logical relationships between features in input time-series sequences using only two dendrite-inspired neuron groups, effectively capturing temporal dependencies. This work represents the first successful extension of dendrite-like neurons from static data processing to dynamic temporal modeling.Systematic experiments on 18 UCR time-series classification datasets and 9 time-series regression datasets, along with comprehensive comparisons with mainstream models including LSTM, GRU, BiLSTM, TCN, and Transformer, demonstrate that Memory-DD achieves an average accuracy of 89.41\% in classification tasks while using only 50\% of the parameters of a conventional LSTM and reducing computational complexity by 27.7\%. In regression tasks, Memory-DD reaches predictive performance comparable to LSTM. Ablation studies further confirm the effectiveness of core components such as the Hadamard product, residual connections, and weight sharing, providing both theoretical and empirical support for lightweight temporal modeling.The success of Memory-DD not only fills a technological gap for dendrite-like neurons in the temporal prediction domain but also establishes a new research paradigm for building efficient and biologically plausible AI systems. It holds significant application potential and development prospects in resource-constrained scenarios, such as edge computing and real-time prediction.

\newpage

\vfill

\end{document}